\documentclass[lettersize,journal]{IEEEtran}
\usepackage{amsmath,amsfonts}
\usepackage{algpseudocode}
\usepackage{algorithmicx}
\usepackage{algorithm}
\usepackage{array}
\usepackage[caption=false,font=normalsize,labelfont=sf,textfont=sf]{subfig}
\usepackage{textcomp}
\usepackage{stfloats}
\usepackage{url}
\usepackage{verbatim}
\usepackage{booktabs}
\usepackage{multirow}
\usepackage{makecell}
\usepackage{colortbl}
\usepackage{xcolor}
\usepackage{graphicx}
\usepackage{cite}
\usepackage{hyperref}

\newcolumntype{G}{>{\columncolor{gray!12}}c}
\newcommand{\ours}{\textsc{VECTOR-Drive}}

\hypersetup{
    colorlinks=true,
    linkcolor=blue,
    citecolor=blue,
    filecolor=magenta,
    urlcolor=cyan,
}

\begin{document}
\onecolumn
\clearpage
\twocolumn

\title{VECTOR-Drive: Tightly Coupled Vision-Language and Trajectory Expert Routing for End-to-End Autonomous Driving}

\author{
\textbf{Rui Zhao}$^{1,2}$, \textbf{Jianlin Yu}$^{1,2}$, \textbf{Zhenhai Gao}$^{1,2}$, \textbf{Jiaqiao Liu}$^{3}$, and \textbf{Gao Fei}$^{1,2}$\\[1.2em]
$^{1}$College of Automotive Engineering, Jilin University,\\[0.15em]
$^{2}$The National Key Laboratory of Automotive Chassis Integration and Bionics, Jilin University,\\[0.15em]
$^{3}$ReeFocus AI Technology
}

\markboth{IEEE Robotics and Automation Letters}{Anonymous Author(s): VECTOR-Drive}

\maketitle

\begin{abstract}
End-to-end autonomous driving requires models to understand traffic scenes, infer driving intent, and generate executable motion plans. Recent vision-language-action (VLA) models inherit semantic priors from large-scale vision-language pretraining, yet still face a coupling trade-off: fully shared backbones preserve multimodal interaction but may entangle language reasoning and trajectory prediction, whereas decoupled reasoning-action pipelines reduce task conflict but weaken semantic-motion coupling. We propose \textbf{\ours}, a tightly coupled VLA framework built on Qwen2.5-VL-3B. \ours\ keeps all tokens coupled through shared self-attention and routes feed-forward computation according to token semantics. Vision and language tokens are processed by a Vision-Language Expert to preserve semantic priors, while target-point, ego-state, and noisy action tokens are routed to a Trajectory Expert for motion-specific computation. On the action-token pathway, a flow-matching planner refines noisy action tokens into future waypoints and speed profiles. This design couples semantic reasoning and motion planning within a single multimodal Transformer while separating task-specific FFN computation. On Bench2Drive, \ours\ achieves 88.91 Driving Score and outperforms representative end-to-end and VLA-based baselines. Qualitative results and ablations further validate the benefits of shared attention, semantic-aware expert routing, progressive training, and flow-based action decoding.The project page is available at: \url{https://github.com/InfiniDrive/VECTOR-Drive}.
\end{abstract}

\begin{IEEEkeywords}
End-to-end autonomous driving, vision-language-action model, trajectory planning, expert routing, flow matching.
\end{IEEEkeywords}

\begin{figure*}[!t]
\centering
\includegraphics[width=0.86\textwidth]{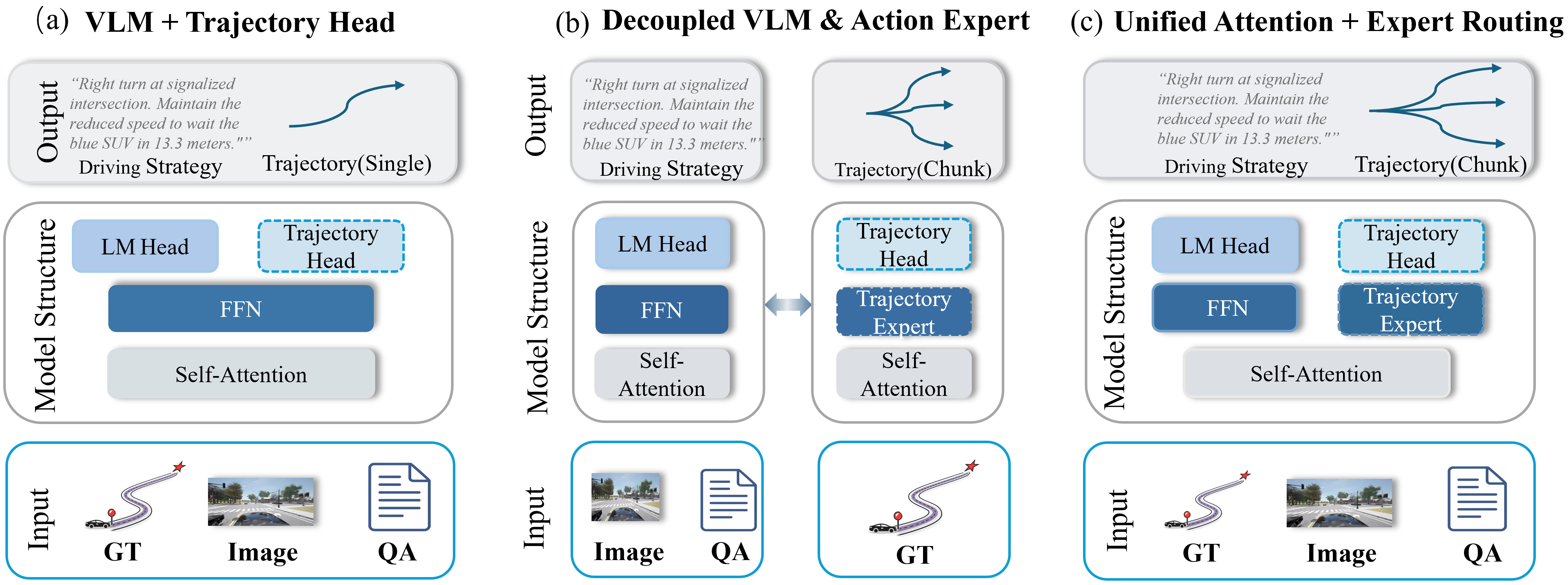}
\caption{\textbf{Three VLA design paradigms.} Left: a shared VLM predicts actions with a single trajectory head. Middle: reasoning and chunk-level motion generation are separated. Right: our shared-attention and expert-routed design preserves multimodal interaction while routing motion-related computation to a dedicated Trajectory Expert.}
\label{fig1}
\vspace{-5pt}
\end{figure*}

\section{Introduction}

\IEEEPARstart{E}{nd-to-end} autonomous driving maps multimodal observations directly to future trajectories, enabling perception, prediction, and planning to be optimized in a unified framework~\cite{prakash2021multi,hu2023planning,weng2024drive,liao2025diffusiondrive,li2024enhancing,li2024hydra}. Recent methods have advanced this paradigm through trajectory-guided control, scalable planning decoders, adapter-based perception-planning decoupling, vectorized scene representations, and unified transformer architectures~\cite{wu2022tcp,jia2023thinktwice,jia2023driveadapter,jiang2023vad,jia2025drivetransformer}. Despite progress on closed-loop benchmarks such as Bench2Drive~\cite{jia2024bench2drive,liao2025diffusiondrive}, long-tail interactions and environmental shifts can still cause accumulated planning errors~\cite{sima2024drivelm,jiang2024senna}.

Vision--Language--Action (VLA) models address this challenge by exploiting large-scale vision-language pretraining and semantic supervision~\cite{zitkovich2023rt,sima2024drivelm,tian2024drivevlm,renz2025simlingo}. Recent driving VLA systems further introduce language-action alignment, vision-language instructed action generation, and adaptive reasoning to improve semantic grounding and decision robustness~\cite{renz2025simlingo,fu2025orion,zhou2025autovla}. A common design adapts a pretrained VLM or Large Language Model (LLM) to driving by attaching a trajectory or control head to the shared backbone, as illustrated in Fig.~\ref{fig1}(a)~\cite{sima2024drivelm,renz2025simlingo,wang2024drivecot,jiang2024senna,zhao2025sce2drivex}. While this preserves multimodal interaction, the same transformer and FFN parameters must support both discrete autoregressive reasoning and continuous geometric prediction. Due to their different inductive biases and optimization behavior, fully shared computation may introduce task interference and degrade motion accuracy~\cite{yu2020gradient,ding2023mitigating}.

A complementary direction separates high-level reasoning from low-level action generation, as shown in Fig.~\ref{fig1}(b)~\cite{jiang2024senna,black2024pi_0,li2025drivevlaw0,yang2025drivemoe}. In such pipelines, a VLM provides semantic representations, while a separate action module or trajectory expert predicts chunk-level motion~\cite{zitkovich2023rt,black2024pi_0}. Although this reduces optimization pressure on the shared backbone, trajectory generation often depends on semantics through a coarse conditioning interface~\cite{renz2025simlingo,li2026drive}. This can limit interactive driving, where semantic constraints must be transferred to fine-grained, dynamically feasible trajectories.

We propose \textbf{\ours} , a tightly coupled VLA  framework Fig.~\ref{fig1}(c) for end-to-end autonomous driving. Instead of fully sharing or fully separating reasoning and action generation, \ours\ shares self-attention across all tokens and routes FFN computation according to token semantics. Vision and language tokens are processed by a Vision-Language Expert, while target-point, ego-state, and noisy action tokens are routed to a Trajectory Expert. On the action-token pathway, a flow-matching planner transforms noisy action representations into future waypoints and speed profiles. Thus, semantic reasoning and trajectory generation remain coupled through attention, while language-oriented and motion-oriented computation is separated in FFN layers.

We evaluate \textbf{\ours} on Bench2Drive in CARLA. It achieves 88.91 Driving Score and 71.82\% Success Rate. Ablations further isolate the effects of shared attention, semantic-aware FFN routing, progressive training, and flow-based trajectory decoding.

The main contributions are:
\begin{itemize}
    \item We propose a unified VLA architecture that combines \textbf{shared self-attention} for cross-modal interaction with \textbf{semantic-aware expert routing} for task-specific FFN computation, reducing interference between language reasoning and motion planning.
    \item We design a flow-matching planner on the routed action-token pathway to refine noisy action tokens into future waypoints and temporal speed profiles under multimodal conditioning.
    \item Experiments on Bench2Drive show that \textbf{\ours} outperforms representative end-to-end and VLA-based baselines, with ablations validating expert routing, three-stage training, and flow-based action decoding.
\end{itemize}

\section{Related Work}
\label{sec:related}

\subsection{Vision-Language-Action Models for Autonomous Driving}
\label{sec:rw_vla}

Large-scale vision-language pretraining has promoted VLA models for autonomous driving. Existing methods typically adapt a pretrained VLM or LLM by jointly encoding visual observations, navigation cues, ego states, and language prompts, followed by a trajectory or control head~\cite{zitkovich2023rt,sima2024drivelm,tian2024drivevlm,renz2025simlingo}. Recent works further improve driving-oriented semantic representations. FLARE learns future-aware latent representations from VLM features, while ReCogDrive combines VLM-based driving cognition with a diffusion planner for reasoning-guided planning~\cite{xie2026flare,li2025recogdrive}. These studies demonstrate the value of VLM priors for long-tail scene understanding and interpretable driving decisions.

However, most VLA driving models still rely on largely shared computation for language reasoning and motion prediction. Although shared backbones preserve multimodal interaction, the same transformer and FFN parameters must support discrete language generation and continuous trajectory regression. Since these objectives differ in optimization behavior, fully shared computation may introduce task interference and weaken motion accuracy~\cite{yu2020gradient,ding2023mitigating}. Moreover, when semantic reasoning is connected to planning only through text, latent summaries, or high-level decisions, fine-grained trajectory generation is only indirectly conditioned on scene understanding, limiting reasoning-control coupling in interactive driving.

\subsection{Decoupled Action Generation and Continuous Planning}
\label{sec:rw_action}

Recent works decouple action generation from vision-language reasoning to improve continuous control. In robotics, $\pi_0$ introduces a flow-matching action model on top of a pretrained vision-language backbone~\cite{black2024pi_0}. This paradigm has also been explored in autonomous driving. DriveVLA-W0 introduces a lightweight action expert, while DriveMoE extends the $\pi_0$-style formulation with mixture-of-experts modules for scene-specialized perception and skill-specialized action generation~\cite{li2025drivevlaw0,yang2025drivemoe}. These designs reduce the burden on general-purpose VLM representations and strengthen action modeling.

Diffusion- and flow-based planners further improve continuous trajectory generation by modeling complex future distributions beyond deterministic regression~\cite{liao2025diffusiondrive,black2024pi_0,li2025recogdrive}. However, fully separated reasoning-action pipelines often reduce the VLM to a conditioning module, weakening bidirectional interaction between semantic understanding and motion generation.

In contrast, our method keeps all modalities in a unified token sequence. Shared self-attention preserves cross-modal interaction at each layer, while token-type-aware FFN routing separates vision-language and trajectory-specific computation. Semantic reasoning and motion planning therefore remain tightly coupled through attention, while their FFN transformations are decoupled to reduce task interference.

\section{Method}
\label{sec:method}

\begin{figure*}[t]
\centering
\includegraphics[width=0.86\textwidth]{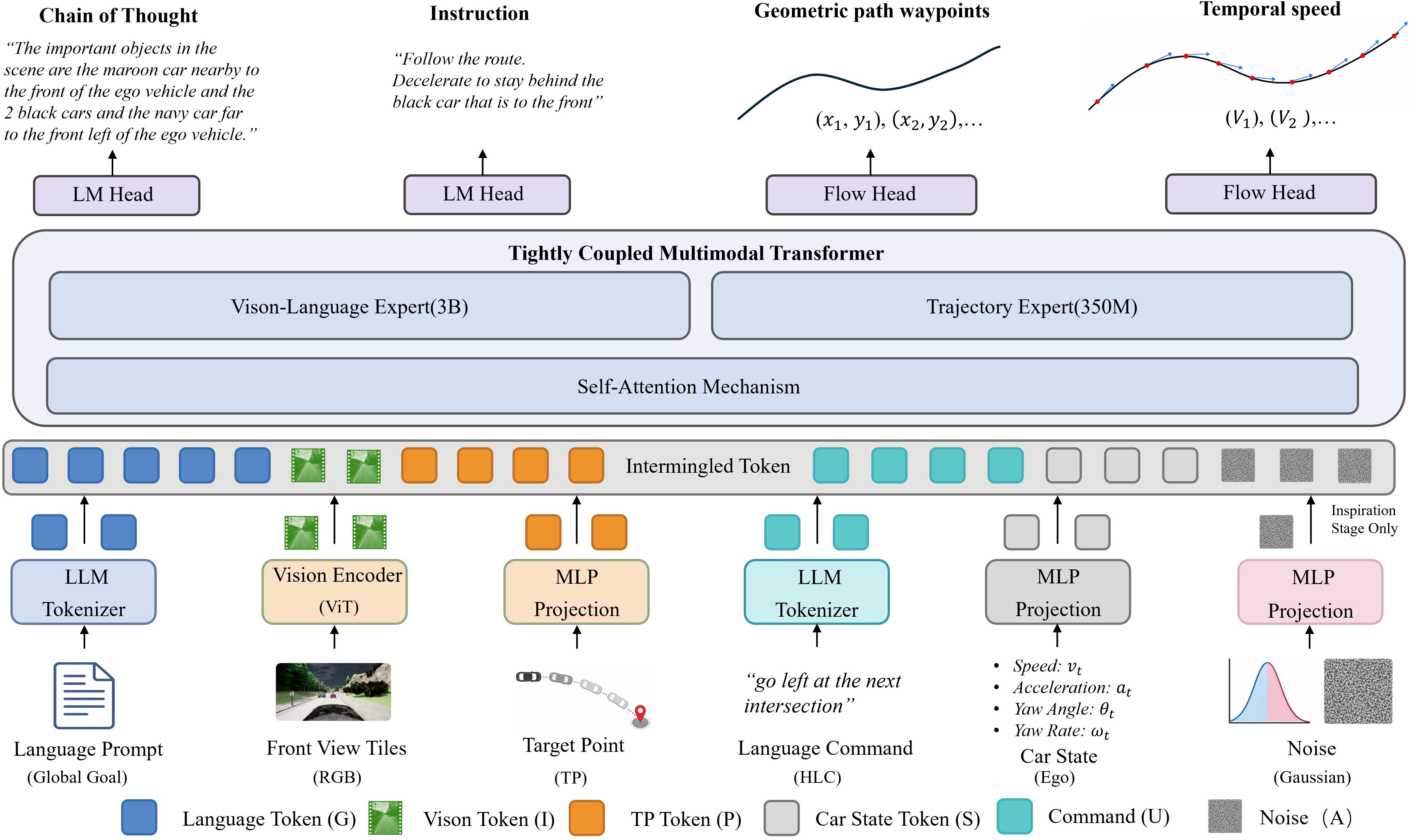}
\caption{\textbf{Overall architecture of \ours.} Visual observations, navigation conditions, language commands, ego states, and noisy action states are organized as an interleaved multimodal token sequence. Shared self-attention preserves cross-modal interaction, while semantic-aware FFN routing separates vision-language and trajectory-oriented computation.}
\label{fig:overall_arch}
\vspace{-4mm}
\end{figure*}

We formulate end-to-end driving as unified multimodal reasoning and motion generation. Built on Qwen2.5-VL-3B, the model maps visual observations, navigation conditions, language commands, ego states, and noisy action representations into an interleaved token sequence. A tightly coupled multimodal Transformer processes this sequence with shared self-attention and semantic-aware FFN routing. The LM Head generates chain-of-thought reasoning and driving instructions, while the Flow Head predicts future waypoints and temporal speed profiles. Fig.~\ref{fig:overall_arch} illustrates the architecture of \ours.

\subsection{Problem Formulation}
\label{subsec:problem_formulation}

We formulate end-to-end driving as conditional multimodal generation. Given a front-view RGB frame $I$, a global-goal prompt $G$, a target point $P$, a high-level command $U$, and an ego state $S$, the model jointly produces language-level reasoning and motion outputs. The ego state is
\begin{equation}
S=[v, a, \theta, \dot{\theta}],
\label{eq:ego_state}
\end{equation}
where $v$, $a$, $\theta$, and $\dot{\theta}$ denote ego speed, acceleration, yaw angle, and yaw rate.

The language output consists of chain-of-thought reasoning and driving instruction text:
\begin{equation}
O=\{o_1,\ldots,o_K\}.
\label{eq:language_output}
\end{equation}
The motion output contains a spatial path and a temporal speed profile:
\begin{equation}
Y^{p}=\{p_1,\ldots,p_{20}\}, \qquad p_k=(x_k,y_k)\in\mathbb{R}^{2},
\label{eq:path_output}
\end{equation}
where one waypoint is predicted per meter over a $20$-meter horizon, and
\begin{equation}
Y^{v}=\{v_1,\ldots,v_{10}\}, \qquad v_q\in\mathbb{R},
\label{eq:speed_output}
\end{equation}
where $v_q$ is the speed at the $q$-th future step.

The conditional mapping is
\begin{equation}
(O,Y^{p},Y^{v})=\mathcal{F}(I,G,P,U,S),
\label{eq:overall_mapping}
\end{equation}
where $\mathcal{F}$ denotes the proposed multimodal driving model. The language output describes scene-level reasoning and driving intent, while $Y^{p}$ and $Y^{v}$ provide the spatial path and temporal speed profile for downstream control.

\subsection{Unified Tokenization}
\label{subsec:unified_tokenization}

All modalities are mapped into a unified token space. The global-goal prompt $G$ and high-level command $U$ are tokenized into $Z_G$ and $Z_U$. The front-view image $I$ is encoded into visual tokens $Z_I$. The target point $P$ and ego state $S=[v,a,\theta,\dot{\theta}]$ are projected into the shared hidden space by lightweight MLPs, producing $Z_P$ and $Z_S$.

For flow-based motion generation, a noisy action representation is introduced. Let $X_{\tau}$ be the noisy path state at interpolation time $\tau$, and let $e(\tau)$ be its time embedding. The noisy action tokens are
\begin{equation}
Z_A(\tau)=\phi(X_{\tau},e(\tau)),
\label{eq:action_tokens}
\end{equation}
where $\phi(\cdot)$ denotes the action-token preprocessor.

All token groups are arranged into an interleaved multimodal sequence:
\begin{equation}
Z=\mathrm{Interleave}
\left(
Z_G,Z_I,Z_P,Z_U,Z_S,Z_A
\right),
\label{eq:unified_sequence}
\end{equation}
where the order follows Fig.~\ref{fig:overall_arch}, and $z_i$ denotes the $i$-th token. This unified sequence enables cross-modal interaction before task-specific decoding.

\subsection{Semantic-Aware Expert Routing}
\label{subsec:routing}

The interleaved sequence $Z$ is used as the initial hidden sequence, $\mathbf{H}^{(0)}=Z$. Each Transformer layer contains shared self-attention and two expert-specific FFN branches. Routing is applied after attention: all tokens first exchange context through shared self-attention, and the contextualized states are then routed according to token semantics. This preserves semantic-to-motion interaction while separating task-specific feed-forward computation.

Given $\mathbf{H}^{(l-1)}\in\mathbb{R}^{N\times d}$ at layer
$l\in\{1,\ldots,L\}$, shared self-attention is applied to the full sequence:
\begin{equation}
\tilde{\mathbf{H}}^{(l)}
=
\mathbf{H}^{(l-1)}
+
\mathrm{Attn}^{(l)}
\left(
\mathrm{LN}\left(\mathbf{H}^{(l-1)}\right)
\right).
\label{eq:shared_attn}
\end{equation}
where $\mathrm{LN}(\cdot)$ denotes layer normalization. The attention operator uses shared $Q$, $K$, $V$, and $O$ projections:
\begin{equation}
\mathrm{Attn}^{(l)}(\mathbf{X})
=
\mathrm{Softmax}
\left(
\frac{\mathbf{Q}\mathbf{K}^{\top}}{\sqrt{d}}
+
\mathbf{M}
\right)
\mathbf{V}\mathbf{W}_{O}^{(l)}.
\label{eq:attn_detail}
\end{equation}
where $\mathbf{Q}=\mathbf{X}\mathbf{W}_{Q}^{(l)}$, $\mathbf{K}=\mathbf{X}\mathbf{W}_{K}^{(l)}$, and $\mathbf{V}=\mathbf{X}\mathbf{W}_{V}^{(l)}$. Thus, action tokens aggregate semantic and driving-condition context at each layer.

After shared attention, let $\tilde{\mathbf{h}}_{i}^{(l)}$ denote the
post-attention hidden state of the $i$-th token in
$Z=\{z_1,\ldots,z_N\}$. We define
$\mathcal{I}_{\mathrm{vl}}=\{i\mid z_i\in Z_G\cup Z_I\cup Z_U\}$ and
$\mathcal{I}_{\mathrm{act}}=\{i\mid z_i\in Z_P\cup Z_S\cup Z_A\}$.
The routed hidden sequences are obtained by ordered gathering:
\begin{equation}
\tilde{\mathbf{H}}_{\mathrm{vl}}^{(l)}
=
\mathcal{C}_{i\in\mathcal{I}_{\mathrm{vl}}}
\left(\tilde{\mathbf{h}}_{i}^{(l)}\right),
\label{eq:z_vl}
\end{equation}
\begin{equation}
\tilde{\mathbf{H}}_{\mathrm{act}}^{(l)}
=
\mathcal{C}_{i\in\mathcal{I}_{\mathrm{act}}}
\left(\tilde{\mathbf{h}}_{i}^{(l)}\right).
\label{eq:z_act}
\end{equation}
where $\mathcal{C}(\cdot)$ denotes ordered concatenation along the token dimension following the original order in $Z$.

The routed expert inputs are
\begin{equation}
\mathbf{O}_{1}^{(l)}
=
\mathrm{LN}\!\left(\tilde{\mathbf{H}}_{\mathrm{vl}}^{(l)}\right),
\qquad
\mathbf{O}_{2}^{(l)}
=
\mathrm{LN}\!\left(\tilde{\mathbf{H}}_{\mathrm{act}}^{(l)}\right),
\label{eq:o1o2_input}
\end{equation}
corresponding to the Vision-Language Expert and the Trajectory Expert.

Each expert uses a gated FFN with gate, up, and down projections:
\begin{equation}
\mathrm{FFN}_{e}^{(l)}(\mathbf{X})
=
\left[
\sigma\!\left(\mathbf{X}\mathbf{W}_{e,\mathrm{gate}}^{(l)}\right)
\odot
\left(\mathbf{X}\mathbf{W}_{e,\mathrm{up}}^{(l)}\right)
\right]
\mathbf{W}_{e,\mathrm{down}}^{(l)},
\label{eq:expert_internal}
\end{equation}
where $e$ indexes the expert branch, with $e=0$ for the Vision-Language
Expert and $e=1$ for the Trajectory Expert. Here, $\sigma(\cdot)$ is SiLU,
and $\odot$ denotes element-wise multiplication.

The expert outputs are
\begin{equation}
\mathbf{O}_{1y}^{(l)}
=
\mathrm{FFN}_{0}^{(l)}(\mathbf{O}_{1}^{(l)}),
\qquad
\mathbf{O}_{2y}^{(l)}
=
\mathrm{FFN}_{1}^{(l)}(\mathbf{O}_{2}^{(l)}).
\label{eq:o1y_o2y}
\end{equation}
They are then merged back into the original token order:
\begin{equation}
\mathbf{H}^{(l)}
=
\mathrm{Merge}_{Z}
\left(
\tilde{\mathbf{H}}_{\mathrm{vl}}^{(l)}
+
\mathbf{O}_{1y}^{(l)},
\;
\tilde{\mathbf{H}}_{\mathrm{act}}^{(l)}
+
\mathbf{O}_{2y}^{(l)}
\right).
\label{eq:merge_back}
\end{equation}
where $\mathrm{Merge}_{Z}(\cdot)$ restores the interleaved order.

This attention--routing--merge process repeats at every layer. After merging, motion tokens remain conditioned on refined semantic features, while the causal mask prevents action-to-language information leakage. Thus, shared attention conditions motion generation on semantics, while separate FFNs reduce task interference. The routing is static and token-type-aware rather than learned top-$k$ routing, avoiding additional router parameters and load-balancing losses.

To improve waypoint-level consistency while preventing action leakage into language representations, we use a causal multimodal attention mask and allow bidirectional attention only within noisy action tokens:
\begin{equation}
M_{ij}=
\begin{cases}
0, & z_i,z_j\in Z_A,\\
M_{\mathrm{causal}}(i,j), & \text{otherwise}.
\end{cases}
\label{eq:mask}
\end{equation}
With $Z_A$ placed at the end of the sequence, action tokens can attend to preceding semantic and driving-condition tokens, but language-side tokens cannot access future action states.

\subsection{Generation of Chain-of-Thought and Driving Instruction}
\label{subsec:language_generation}

The language branch generates chain-of-thought reasoning and driving instruction
text from the final routed Vision-Language hidden states
$\mathbf{H}_{\mathrm{vl}}^{(L)}$. The LM Head performs autoregressive decoding:
\begin{equation}
p(O\mid \mathbf{H}_{\mathrm{vl}}^{(L)})
=
\prod_{k=1}^{K}
p(o_k\mid o_{<k},\mathbf{H}_{\mathrm{vl}}^{(L)}).
\label{eq:lm_factorization}
\end{equation}

Language decoding is performed only from Vision-Language hidden states, while
alignment with motion planning is encouraged by the shared multimodal context and
joint optimization. Under the causal mask, action tokens attend to semantic and
driving-condition tokens, enabling semantic-conditioned trajectory generation
without future-action leakage into language decoding.

\subsection{Flow-Matching Action Planner with Multimodal Conditioning}
\label{subsec:flow_matching}

Fine-grained motion is generated by a flow-matching planner for both spatial
path prediction and temporal speed profiling. The spatial branch models future
route geometry, and the speed branch captures temporal motion evolution. We
detail the spatial formulation and apply the same flow-matching principle to
speed generation.

\noindent\textbf{Spatial path generation.}
The spatial branch predicts a $20$-meter future path with $20$ ordered waypoints. Let $Y^{p}\in\mathbb{R}^{20\times 2}$ denote the ground-truth path, and let $\boldsymbol{\epsilon}\sim\mathcal{N}(0,I)$ be Gaussian noise with the same shape. For $\tau\in[0,1]$, the noisy path state is
\begin{equation}
X_{\tau}=(1-\tau)\boldsymbol{\epsilon}+\tau Y^{p},
\label{eq:xt}
\end{equation}
with the target vector field
\begin{equation}
\mathbf{v}^{\star}=Y^{p}-\boldsymbol{\epsilon}.
\label{eq:v_star}
\end{equation}

\begin{figure}[t]
\centering
\includegraphics[width=\columnwidth]{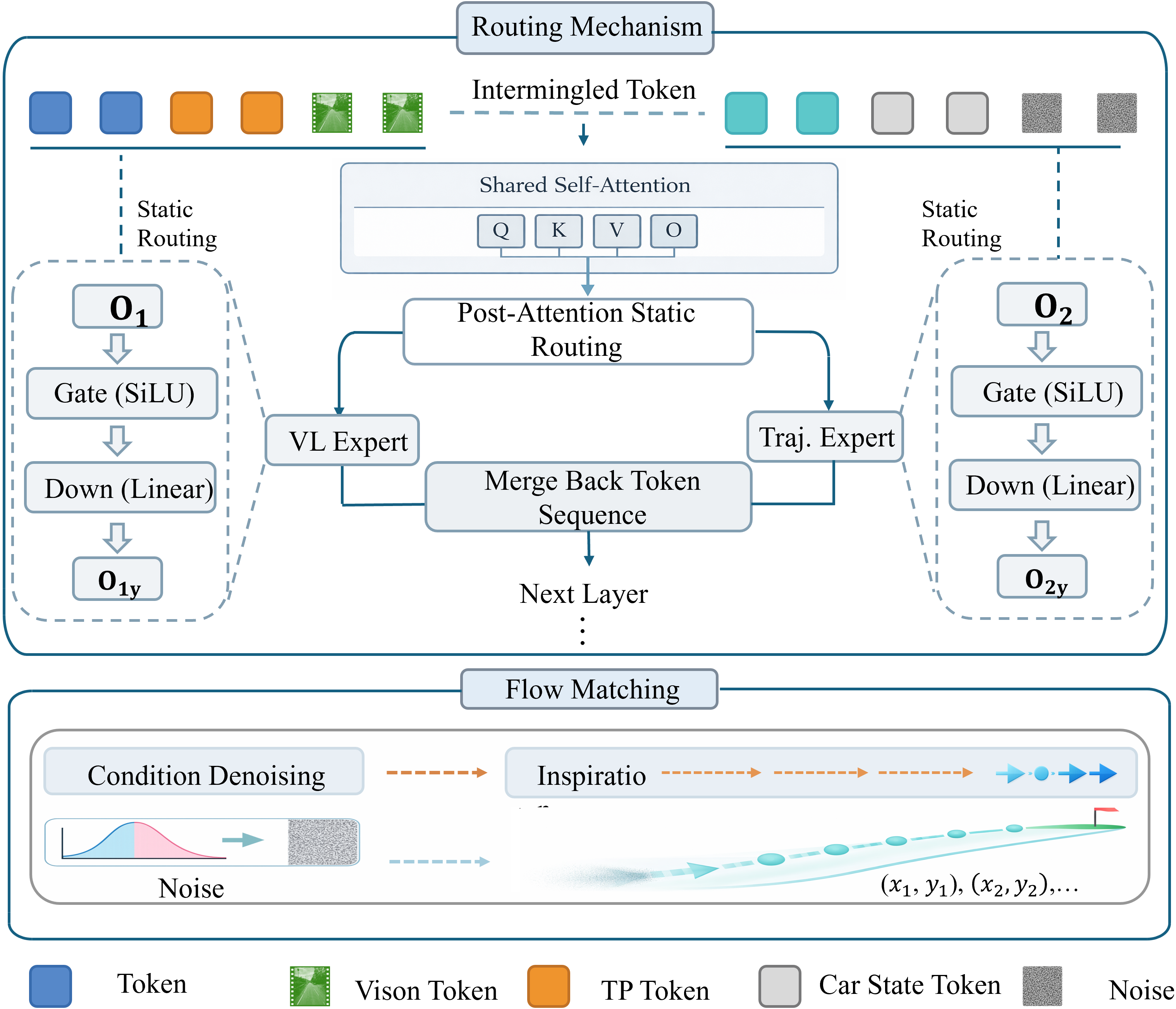}
\caption{\textbf{Routing and planning.} 
\textbf{Top:} Interleaved tokens are processed by shared self-attention, routed to the Vision-Language or Trajectory Expert, and merged back into the sequence. 
\textbf{Bottom:} The flow planner refines Gaussian noise into future trajectories under semantic and driving-condition guidance.}
\label{fig:router_flow}
\vspace{-1mm}
\end{figure}

At each $\tau$, $X_{\tau}$ and its time embedding $e(\tau)$ are encoded into
noisy action tokens $Z_A(\tau)$ by Eq.~\eqref{eq:action_tokens}, which replace
$Z_A$ at the corresponding action-token positions in the interleaved sequence.
After shared attention, expert routing, and merge-back operations, the final
action-token hidden states are obtained by ordered gathering:
\begin{equation}
\mathbf{H}_{A}^{(L)}(\tau)
=
\mathcal{C}_{i\in\mathcal{I}_{A}}
\left(
\mathbf{h}_{i}^{(L)}(\tau)
\right),
\qquad
\mathcal{I}_{A}=\{i\mid z_i\in Z_A\}.
\label{eq:final_action_hidden}
\end{equation}
Thus, $Z_A(\tau)$ provides the current noisy state, while
$\mathbf{H}_{A}^{(L)}(\tau)$ provides the semantic-conditioned motion
representation used by the Flow Head.

The implicit semantic condition is obtained from the final Vision-Language hidden states:
\begin{equation}
\mathbf{C}_{\mathrm{imp}}^{(L)}
=
\psi_{\mathrm{vl}}
\left(
\mathbf{H}_{\mathrm{vl}}^{(L)}
\right),
\label{eq:implicit_condition}
\end{equation}
where $\psi_{\mathrm{vl}}(\cdot)$ projects semantic hidden states into the
flow-head input space. The explicit driving condition is formed by projecting
ego state, navigation representation, and target point:
\begin{equation}
\mathbf{C}_{\mathrm{exp}}
=
\mathcal{C}
\left(
\psi_{S}(S),
\psi_{\mathrm{nav}}(\bar{z}_{\mathrm{nav}}),
\psi_{P}(P)
\right),
\label{eq:explicit_condition}
\end{equation}
where
\begin{equation}
\bar{z}_{\mathrm{nav}}
=
\mathrm{Pool}
\left(
\mathcal{C}(Z_G,Z_U)
\right).
\label{eq:nav_pool}
\end{equation}

The complete flow condition is
\begin{equation}
\mathbf{C}_{\mathrm{fm}}^{(L)}
=
\mathcal{C}
\left(
\mathbf{C}_{\mathrm{imp}}^{(L)},
\mathbf{C}_{\mathrm{exp}}
\right).
\label{eq:flow_condition}
\end{equation}
The Flow Head predicts the vector field from the routed action hidden states and the multimodal condition:
\begin{equation}
\mathbf{F}_{\tau}
=
\mathcal{C}
\left(
\psi_{A}
\left(
\mathbf{H}_{A}^{(L)}(\tau)
\right),
\mathbf{C}_{\mathrm{fm}}^{(L)}
\right),
\label{eq:flow_input}
\end{equation}
\begin{equation}
\hat{\mathbf{v}}_{\omega}
\left(
X_{\tau},\tau
\mid
\mathbf{C}_{\mathrm{fm}}^{(L)}
\right)
=
f_{\omega}
\left(
\mathbf{F}_{\tau}
\right),
\label{eq:v_hat}
\end{equation}
where $f_{\omega}(\cdot)$ denotes the flow head.

At inference, the path is initialized from Gaussian noise:
\begin{equation}
X_0\sim\mathcal{N}(0,I),
\label{eq:x0}
\end{equation}
and the conditional ODE
\begin{equation}
\frac{dX_{\tau}}{d\tau}
=
\hat{\mathbf{v}}_{\omega}
\left(
X_{\tau},\tau
\mid
\mathbf{C}_{\mathrm{fm}}^{(L)}
\right),
\qquad \tau\in[0,1],
\label{eq:ode}
\end{equation}
is solved to obtain $\hat{Y}^{p}$. During ODE solving, the current state
$X_{\tau}$ is repeatedly converted into $Z_A(\tau)$, processed by the routed
Transformer, and decoded by the Flow Head.

\noindent\textbf{Temporal speed generation.}
The model also predicts a future speed profile:
\begin{equation}
\hat{Y}^{v}=\{\hat{v}_1,\ldots,\hat{v}_{10}\}, \qquad \hat{v}_q\in\mathbb{R}.
\label{eq:speed_pred}
\end{equation}
Speed generation follows the same flow-matching formulation as spatial path generation. It uses the same implicit semantic condition $\mathbf{C}_{\mathrm{imp}}^{(L)}$ and explicit driving condition $\mathbf{C}_{\mathrm{exp}}$, while replacing the noisy path state with a noisy speed-profile state. The corresponding routed speed hidden states are then decoded by a speed-specific Flow Head.

Thus, spatial path and temporal speed generation share a unified multimodal flow-matching scheme, with separate heads for route geometry and temporal motion evolution.

\subsection{Three-Stage Training Strategy}
\label{subsec:training}

We train the framework in three stages to stabilize vision-language reasoning, specialize motion generation, and align both pathways.

\noindent\textbf{Stage I: Vision-Language Learning.}
The Trajectory Expert and motion heads are frozen, while the Vision-Language Expert and LM Head are optimized for scene understanding, semantic reasoning, and instruction generation:
\begin{equation}
\mathcal{L}_{\mathrm{lang}}
=
-\sum_{k=1}^{K}
\log p
\left(
o_k
\mid
o_{<k},
\mathbf{H}_{\mathrm{vl}}^{(L)}
\right),
\label{eq:lang_loss}
\end{equation}
where $\mathbf{H}_{\mathrm{vl}}^{(L)}$ denotes the final routed Vision-Language hidden states.

\noindent\textbf{Stage II: Driving-Skill Learning.}
The Vision-Language Expert is frozen, while the Trajectory Expert, flow heads, and condition projectors are optimized for motion generation. For the spatial path branch, the flow-matching loss is
\begin{equation}
\mathcal{L}_{\mathrm{path}}
=
\mathbb{E}_{Y^{p},\boldsymbol{\epsilon}^{p},\tau}
\left[
\left\|
f_{\omega}^{p}
\left(
\mathbf{F}_{\tau}^{p}
\right)
-
\left(
Y^{p}-\boldsymbol{\epsilon}^{p}
\right)
\right\|_2^2
\right],
\label{eq:path_flow_loss}
\end{equation}
where $\mathbf{F}_{\tau}^{p}$ is formed by concatenating the projected routed
path-token hidden states with $\mathbf{C}^{(L)}_{\mathrm{fm}}$.

We further use a second-order finite-difference regularizer:
\begin{equation}
\mathcal{L}_{\mathrm{smooth}}
=
\frac{1}{18}
\sum_{k=2}^{19}
\left\|
\hat{p}_{k+1}
-
2\hat{p}_{k}
+
\hat{p}_{k-1}
\right\|_2^2.
\label{eq:smooth_loss}
\end{equation}

The speed branch follows the same flow-matching formulation:
\begin{equation}
\mathcal{L}_{\mathrm{speed}}
=
\mathbb{E}_{Y^{v},\boldsymbol{\epsilon}^{v},\tau}
\left[
\left\|
f_{\eta}^{v}
\left(
\mathbf{F}_{\tau}^{v}
\right)
-
\left(
Y^{v}-\boldsymbol{\epsilon}^{v}
\right)
\right\|_2^2
\right],
\label{eq:speed_flow_loss}
\end{equation}
where $\mathbf{F}_{\tau}^{v}$ is constructed analogously by replacing routed
path-token hidden states with routed speed-profile hidden states.

The driving objective is
\begin{equation}
\mathcal{L}_{\mathrm{drive}}
=
\lambda_{\mathrm{path}}\mathcal{L}_{\mathrm{path}}
+
\lambda_{\mathrm{smooth}}\mathcal{L}_{\mathrm{smooth}}
+
\lambda_{\mathrm{speed}}\mathcal{L}_{\mathrm{speed}},
\label{eq:drive_loss}
\end{equation}
where $\lambda_{\mathrm{path}}$, $\lambda_{\mathrm{smooth}}$, and $\lambda_{\mathrm{speed}}$ are balancing coefficients.

\noindent\textbf{Stage III: Joint Full-Parameter Fine-Tuning.}
All experts and prediction heads are unfrozen and optimized end-to-end:
\begin{equation}
\mathcal{L}
=
\mathcal{L}_{\mathrm{lang}}
+
\mathcal{L}_{\mathrm{drive}}.
\label{eq:final_loss}
\end{equation}

This schedule first adapts the language pathway, then specializes the trajectory pathway, and finally aligns semantic reasoning with flow-based motion generation under shared multimodal context.

\section{Experiments}
\label{sec:experiments}

We evaluate \ours\ on closed-loop performance, qualitative behavior, and component-wise ablations. The experiments compare \ours\ with representative end-to-end and VLA-based baselines on Bench2Drive and analyze semantic-aware routing, three-stage training, and flow-based trajectory decoding.

\subsection{Experimental Settings}

\noindent\textbf{Closed-loop driving benchmark.}
We evaluate on Bench2Drive, a CARLA-based closed-loop benchmark with 44 interactive scenarios and 5 routes per scenario, yielding 220 routes under diverse traffic and weather conditions. We report Driving Score (DS), Success Rate (SR), Efficiency, Comfort, and Multi-Ability. Multi-Ability measures fine-grained behaviors including merging, overtaking, braking, give-way, and traffic-sign compliance.

\noindent\textbf{Implementation details.}
The model is built upon Qwen2.5-VL-3B. Both the Vision-Language Expert and Trajectory Expert are initialized from the corresponding pretrained FFN layers of the backbone; the latter is then specialized for motion-oriented computation. The training set is self-collected in CARLA using the privileged rule-based expert PDM-lite, and contains approximately 3 million frames covering diverse traffic interactions, routes, and environmental conditions. Training uses 4$\times$A800 80GB GPUs with a learning rate of $3\times10^{-5}$. The three stages use 10, 12, and 7 epochs, respectively. Loss weights are set to $\lambda_{\mathrm{path}}=1.0$, $\lambda_{\mathrm{smooth}}=0.1$, and $\lambda_{\mathrm{speed}}=1.0$.

For the flow-matching planner, the future path is initialized from Gaussian noise and refined with an Euler ODE solver. During inference, 10 Euler steps produce the final 20-waypoint trajectory, and the speed branch predicts a 10-step future speed profile.

\subsection{Main Closed-loop Driving Results}

\begin{table*}[t]
\centering
\small
\setlength{\tabcolsep}{4.2pt}
\caption{Main results and multi-ability on Bench2Drive~\cite{jia2024bench2drive}. Higher is better for all metrics.}
\label{tab:main_results}
\resizebox{\textwidth}{!}{
\begin{tabular}{lGGcccccccG}
\toprule
\multirow{2}{*}{Method} & \multicolumn{4}{c}{Closed-loop metrics $\uparrow$} & \multicolumn{6}{c}{Multi-Ability (\%) $\uparrow$} \\
\cmidrule(lr){2-5}\cmidrule(lr){6-11}
& DS & SR (\%) & Efficiency & Comfort & Merging & Overtake & Brake & \makecell{Give-\\Way} & \makecell{Traffic-\\Sign} & Mean \\
\midrule
TCP~\cite{wu2022tcp}               & 40.70 & 15.00 & 54.26 & 47.80 & 16.18 & 20.00 & 20.00 & 10.00 &  6.99 & 14.63 \\
TCP-ctrl~\cite{wu2022tcp}          & 30.47 &  7.27 & 55.97 & \textbf{51.51} & 10.29 &  4.44 & 10.00 & 10.00 &  6.45 &  8.23 \\
TCP-traj~\cite{wu2022tcp}          & 59.90 & 30.00 & 76.54 & 18.08 &  8.89 & 24.29 & 51.67 & 40.00 & 46.28 & 34.22 \\
ThinkTwice~\cite{jia2023thinktwice} & 62.44 & 31.23 & 69.33 & 16.22 & 27.38 & 18.42 & 35.82 & \textbf{50.00} & 54.23 & 37.17 \\
DriveAdapter~\cite{jia2023driveadapter} & 64.22 & 33.08 & 70.22 & 16.01 & 28.82 & 26.38 & 48.76 & \textbf{50.00} & 56.43 & 42.08 \\
\midrule
AD-MLP~\cite{zhai2023admlp}         & 18.05 &  0.00 & 48.45 & 22.63 &  0.00 &  0.00 &  0.00 &  0.00 &  4.35 &  0.87 \\
UniAD-Tiny~\cite{hu2023planning}    & 40.73 & 13.18 &123.92 & 47.04 &  8.89 &  9.33 & 20.00 & 20.00 & 15.43 & 14.73 \\
UniAD-Base~\cite{hu2023planning}    & 45.81 & 16.36 &129.21 & 43.58 & 14.10 & 17.78 & 21.67 & 10.00 & 14.21 & 15.55 \\
VAD~\cite{jiang2023vad}             & 42.35 & 15.00 &157.94 & 46.01 &  8.11 & 24.44 & 18.64 & 20.00 & 19.15 & 18.07 \\
DriveTransformer~\cite{jia2025drivetransformer} & 63.46 & 35.01 &100.64 & 20.78 & 17.57 & 35.00 & 48.36 & 40.00 & 52.10 & 38.60 \\
ReCogDrive~\cite{li2025recogdrive}  & 71.36 & 45.45 &138.18 & 17.45 & 29.73 & 20.00 & 69.09 & 20.00 & 71.34 & 42.03 \\
DriveMoE~\cite{yang2025drivemoe}    & 74.22 & 48.64 &175.96 & 15.31 & 34.67 & 40.00 & 65.45 & 40.00 & 59.44 & 47.91 \\
Orion~\cite{fu2025orion}            & 77.74 & 54.62 &151.48 & 17.38 & 25.00 & 71.11 & 78.33 & 30.00 & 69.15 & 54.72 \\
AutoVLA~\cite{zhou2025autovla}      & 78.84 & 57.73 &146.93 & 39.33 &   --  &   --  &   --  &   --  &   --  &   --  \\
SimLingo~\cite{renz2025simlingo}    & 85.07 & 67.27 &\textbf{259.23} & 33.67 & 53.75 & 68.89 & 81.67 & \textbf{50.00} & 82.11 & 67.28 \\
\midrule
\rowcolor{blue!8}
\ours\ (Ours) & \textbf{88.91} & \textbf{71.82} & 252.10 & 34.20
& \textbf{57.50} & \textbf{78.33} & \textbf{91.67} & \textbf{50.00} & \textbf{82.63} & \textbf{72.03} \\
\bottomrule
\end{tabular}}
\end{table*}

As summarized in Table~\ref{tab:main_results}, \ours\ achieves 88.91 DS and
71.82\% SR on Bench2Drive, outperforming SimLingo by 3.84 DS and 4.55
percentage points in SR. It also improves over ReCogDrive and DriveMoE,
representing decoupled and weakly coupled VLA paradigms, respectively. These
results indicate that shared attention with expert-specific FFN routing provides
an effective interface for transferring scene-level semantics into executable
motion plans.

\ours\ further achieves the highest Multi-Ability mean score of 72.03\%, with
clear gains in merging, overtaking, braking, and traffic-sign compliance. While
SimLingo shows slightly higher Efficiency, \ours\ performs better in DS, SR, and
behavior-level robustness, suggesting improved closed-loop reliability in
interactive driving scenarios.

\subsection{Qualitative Driving Visualization}

\begin{figure*}[t]
\centering
\includegraphics[width=0.96\textwidth]{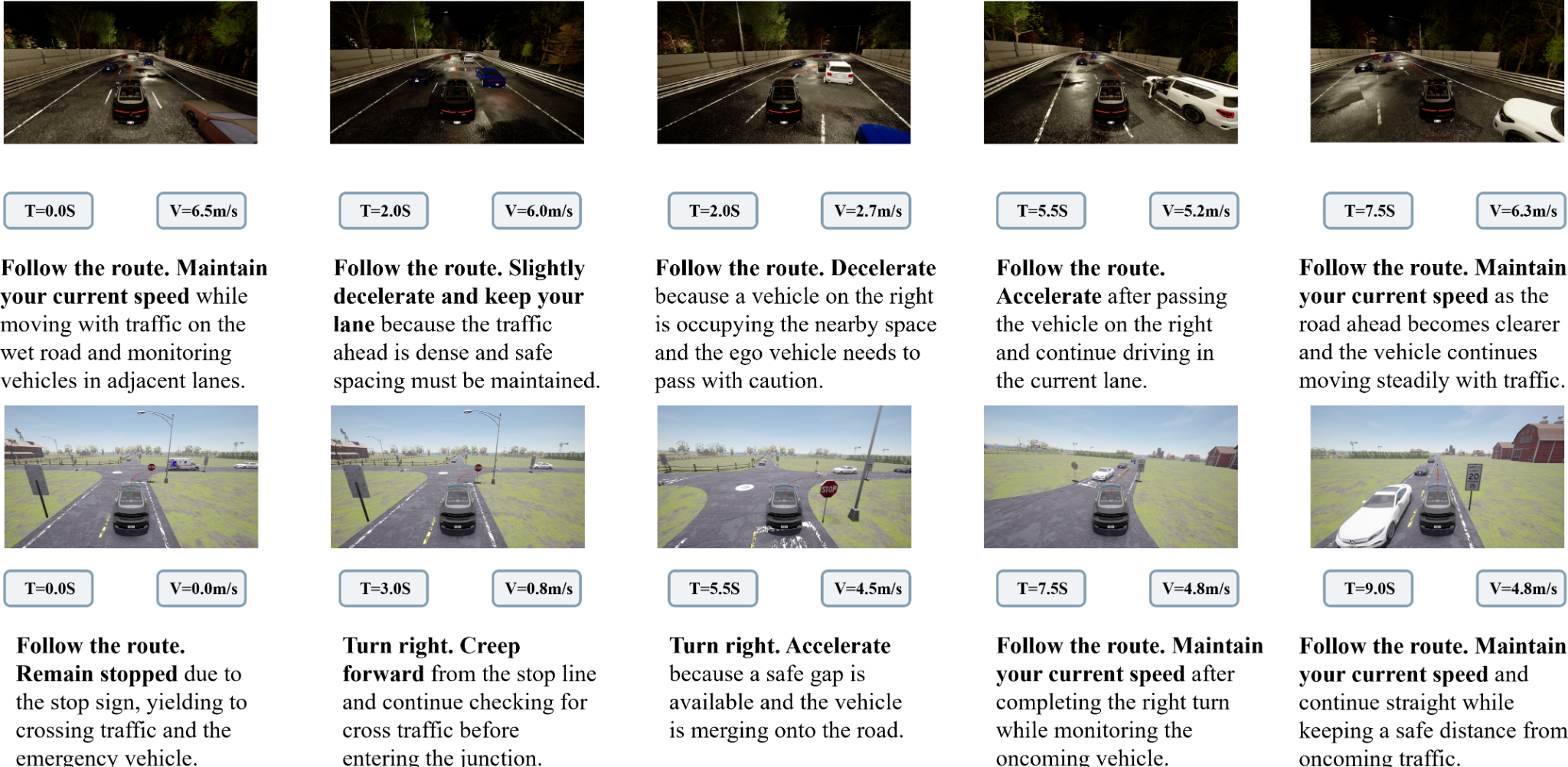}
\caption{\textbf{Qualitative closed-loop visualization.} Two scenarios are shown with time stamps, speed, and language-guided responses. \textbf{Top:} In nighttime wet-road car following, the model maintains speed, decelerates for dense traffic and a nearby right-side vehicle, and accelerates after the interaction resolves. \textbf{Bottom:} At a stop-controlled right turn, it stops, creeps forward to check cross traffic, accelerates after a safe gap, and proceeds steadily.}
\label{fig:qualitative_vis}
\vspace{-4mm}
\end{figure*}

The qualitative cases in Fig.~\ref{fig:qualitative_vis} further illustrate the temporal consistency of
the generated decisions. In the car-following scenario, VECTOR-DRIVE maintains
stable motion when the surrounding traffic is sparse, decelerates as lateral
space becomes constrained by nearby vehicles, and resumes acceleration after the
interaction is resolved. In the stop-controlled right-turn scenario, the model
first yields at the stop line, then performs a cautious creeping behavior to
observe cross traffic, and finally accelerates once a safe gap is available.
These behaviors suggest that the generated language responses and speed profiles
are temporally aligned with the closed-loop driving context.

\subsection{Ablation Studies}

\begin{table}[t]
\centering
\small
\setlength{\tabcolsep}{5pt}
\caption{Ablation on the coupling and routing design. Driving Score (DS) and Success Rate (SR) are reported. Higher is better.}
\label{tab:ablation_routing}
\begin{tabular}{lcc}
\toprule
Method & DS $\uparrow$ & SR (\%) $\uparrow$ \\
\midrule
Decoupled VLM & 83.33 & 64.09 \\
Weakly Coupled VLM & 86.67 & 68.18 \\
Learned Routing & 88.85 & 71.82 \\
Single Action-Oriented Expert & 87.67 & 70.45 \\
\rowcolor{blue!8}
\ours & \textbf{88.91} & \textbf{71.82} \\
\bottomrule
\end{tabular}
\end{table}

To isolate the effect of semantic-motion coupling and expert routing,
Table~\ref{tab:ablation_routing} compares several architectural variants.
Decoupled VLM removes the action expert and uses VLM representations to
condition motion generation, obtaining 83.33 DS and 64.09\% SR. Weakly Coupled
VLM further decouples the shared-attention interaction between vision-language
and action-related tokens, keeping only a coarse semantic conditioning interface;
it improves to 86.67 DS and 68.18\% SR but still lags behind the full model.
These results suggest that robust planning requires both motion-specialized
computation and token-level semantic-motion interaction. Single Action-Oriented
Expert also reduces performance, highlighting the importance of preserving a
dedicated Vision-Language pathway during motion learning.

Learned Routing reaches 88.85 DS and 71.82\% SR, close to the proposed static
routing strategy. This indicates that the main gain comes from shared-attention
coupling and expert-specific FFN computation, rather than from a more complex
learned router. We therefore adopt static semantic-aware routing to avoid
additional router parameters and load-balancing losses.

\begin{table}[t]
\centering
\small
\setlength{\tabcolsep}{5pt}
\caption{Ablation on the three-stage training strategy. Driving Score (DS) and Success Rate (SR) are reported. Higher is better.}
\label{tab:ablation_training}
\begin{tabular}{lcc}
\toprule
Method & DS $\uparrow$ & SR (\%) $\uparrow$ \\
\midrule
Joint training only & 85.67 & 69.09 \\
w/o Stage I VL learning & 87.57 & 70.91 \\
w/o Stage II driving learning & 88.47 & 71.36 \\
w/o Stage III joint fine-tuning & 88.59 & 71.68 \\
\rowcolor{blue!8}
Full three-stage training  & \textbf{88.91} & \textbf{71.82} \\
\bottomrule
\end{tabular}
\end{table}

Table~\ref{tab:ablation_training} evaluates the three-stage training strategy.
Direct joint training reaches 85.67 DS and 69.09\% SR, indicating that jointly
optimizing language reasoning and motion prediction from the start is less
stable. Removing Stage I leads to a larger drop than removing Stage II or Stage
III, suggesting that early vision-language adaptation provides the key semantic
priors for motion learning, while the later stages mainly offer additional refinement.

\begin{table}[t]
\centering
\small
\setlength{\tabcolsep}{5pt}
\caption{Ablation on the motion decoding head. Driving Score (DS) and Success Rate (SR) are reported. Higher is better.}
\label{tab:ablation_motion}
\begin{tabular}{lcc}
\toprule
Method & DS $\uparrow$ & SR (\%) $\uparrow$ \\
\midrule
VLM+Regression & 86.01 & 67.27 \\
VLM+DDPM & 87.67 & 70.45 \\
\rowcolor{blue!8}
\ours & \textbf{88.91} & \textbf{71.82} \\
\bottomrule
\end{tabular}
\end{table}

Finally, Table~\ref{tab:ablation_motion} studies the choice of motion decoder
under the same backbone. The regression decoder obtains 86.01 DS and 67.27\%
SR, while the DDPM decoder improves the result to 87.67 DS and 70.45\% SR.
The proposed flow-matching planner achieves 88.91 DS and 71.82\% SR,
demonstrating that continuous vector-field decoding is more effective for
refining noisy action tokens into executable trajectories under multimodal
conditioning.

\subsection{CoT and Instruction Visualization}

\begin{figure}[t]
\centering
\includegraphics[width=\columnwidth]{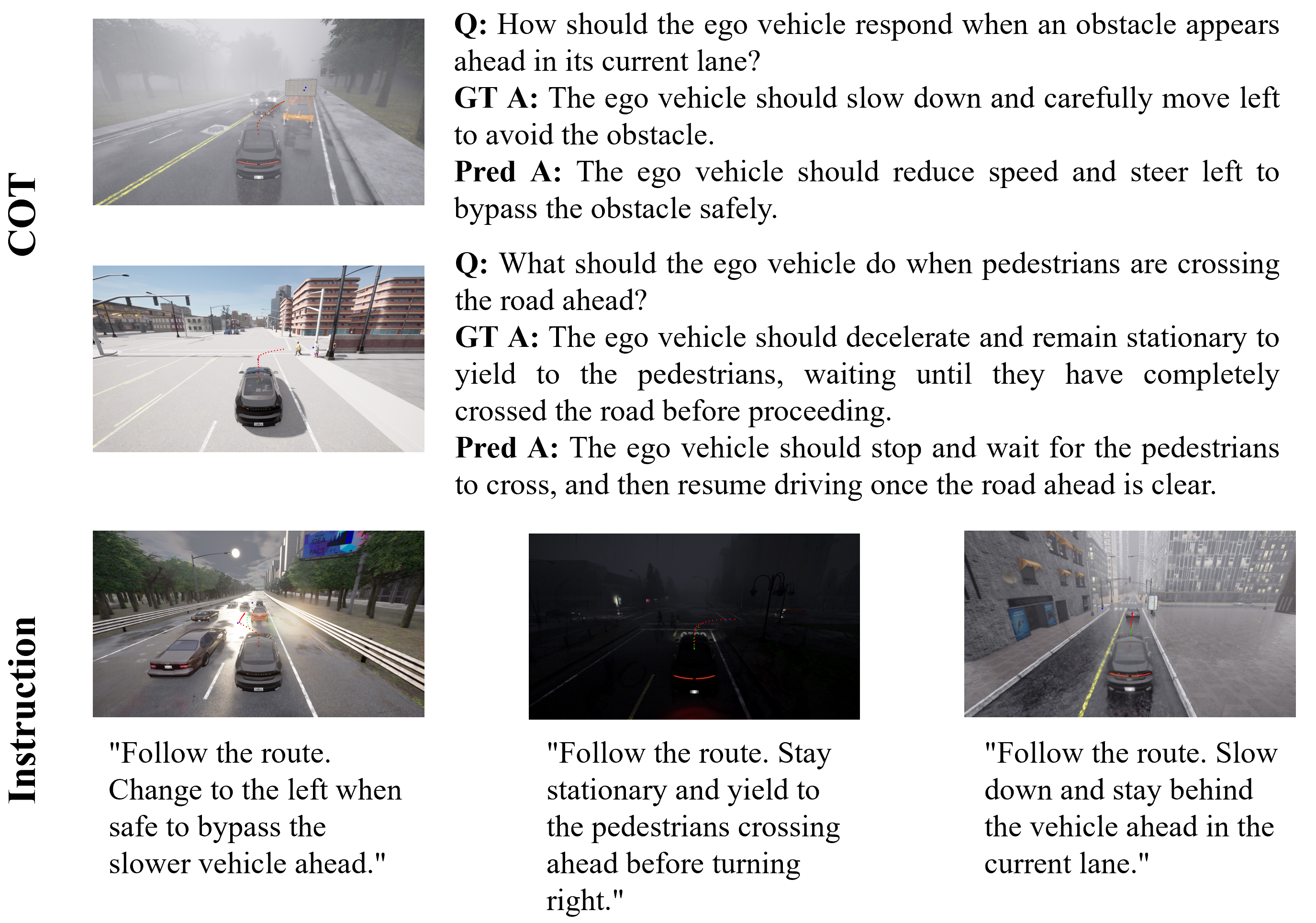}
\caption{\textbf{CoT and instruction visualization.} 
The examples show scene-aware reasoning and concise driving instructions generated by \ours\ under different traffic conditions.}
\label{fig:cot_instruction}
\vspace{-2mm}
\end{figure}

Fig.~\ref{fig:cot_instruction} visualizes the language-generation ability of
\ours. Although language quality is not evaluated as a standalone metric, the
examples show that the model produces scene-aware reasoning and concise
instructions consistent with the intended driving behaviors. Together with the
Stage-I ablation in Table~\ref{tab:ablation_training}, this suggests that
language-side semantic learning provides useful contextual priors for
closed-loop planning.

\subsection{Discussion}

The results indicate that effective VLA driving requires both semantic-motion
coupling and task-specific computation. \ours\ achieves stronger performance in
interaction-heavy scenarios, including merging, overtaking, braking, and
traffic-sign compliance, suggesting that semantic cues are better converted into
executable motion plans. The ablations show that weakening this coupling,
removing the Vision-Language pathway, or replacing the flow planner degrades
closed-loop performance. Overall, shared attention preserves cross-modal
interaction, while expert-specific FFNs reduce interference between language
reasoning and trajectory generation.

\section{Conclusion}
\label{sec:conclusion}

We presented \ours, a tightly coupled VLA framework for end-to-end autonomous driving. By combining shared self-attention, semantic-aware FFN routing, and flow-based action decoding, the model jointly supports scene reasoning and continuous motion generation. Experiments on Bench2Drive demonstrate strong closed-loop performance, while ablations verify the benefits of expert routing, progressive training, and flow-matching planning. These results show that coupling multimodal interaction at the attention level while separating task-specific FFN computation is effective for robust autonomous driving.

\bibliographystyle{IEEEtran}
\bibliography{reference}

@inproceedings{hu2023planning,
  title={Planning-oriented autonomous driving},
  author={Hu, Yihan and Yang, Jiazhi and Chen, Li and Li, Keyu and Sima, Chonghao and Zhu, Xizhou and Chai, Siqi and Du, Senyao and Lin, Tianwei and Wang, Wenhai and others},
  booktitle={Proceedings of the IEEE/CVF conference on computer vision and pattern recognition},
  pages={17853--17862},
  year={2023}
}

@inproceedings{prakash2021multi,
  title={Multi-modal fusion transformer for end-to-end autonomous driving},
  author={Prakash, Aditya and Chitta, Kashyap and Geiger, Andreas},
  booktitle={Proceedings of the IEEE/CVF conference on computer vision and pattern recognition},
  pages={7077--7087},
  year={2021}
}

@inproceedings{weng2024drive,
  title={Para-drive: Parallelized architecture for real-time autonomous driving},
  author={Weng, Xinshuo and Ivanovic, Boris and Wang, Yan and Wang, Yue and Pavone, Marco},
  booktitle={Proceedings of the IEEE/CVF Conference on Computer Vision and Pattern Recognition},
  pages={15449--15458},
  year={2024}
}

@inproceedings{liao2025diffusiondrive,
  title={Diffusiondrive: Truncated diffusion model for end-to-end autonomous driving},
  author={Liao, Bencheng and Chen, Shaoyu and Yin, Haoran and Jiang, Bo and Wang, Cheng and Yan, Sixu and Zhang, Xinbang and Li, Xiangyu and Zhang, Ying and Zhang, Qian and others},
  booktitle={Proceedings of the Computer Vision and Pattern Recognition Conference},
  pages={12037--12047},
  year={2025}
}

@article{li2024hydra,
  title={Hydra-mdp: End-to-end multimodal planning with multi-target hydra-distillation},
  author={Li, Zhenxin and Li, Kailin and Wang, Shihao and Lan, Shiyi and Yu, Zhiding and Ji, Yishen and Li, Zhiqi and Zhu, Ziyue and Kautz, Jan and Wu, Zuxuan and others},
  journal={arXiv preprint arXiv:2406.06978},
  year={2024}
}

@article{li2024enhancing,
  title={Enhancing end-to-end autonomous driving with latent world model},
  author={Li, Yingyan and Fan, Lue and He, Jiawei and Wang, Yuqi and Chen, Yuntao and Zhang, Zhaoxiang and Tan, Tieniu},
  journal={arXiv preprint arXiv:2406.08481},
  year={2024}
}

@inproceedings{zitkovich2023rt,
  title={Rt-2: Vision-language-action models transfer web knowledge to robotic control},
  author={Zitkovich, Brianna and Yu, Tianhe and Xu, Sichun and Xu, Peng and Xiao, Ted and Xia, Fei and Wu, Jialin and Wohlhart, Paul and Welker, Stefan and Wahid, Ayzaan and others},
  booktitle={Conference on Robot Learning},
  pages={2165--2183},
  year={2023},
  organization={PMLR}
}

@article{tian2024drivevlm,
  title={Drivevlm: The convergence of autonomous driving and large vision-language models},
  author={Tian, Xiaoyu and Gu, Junru and Li, Bailin and Liu, Yicheng and Wang, Yang and Zhao, Zhiyong and Zhan, Kun and Jia, Peng and Lang, Xianpeng and Zhao, Hang},
  journal={arXiv preprint arXiv:2402.12289},
  year={2024}
}

@inproceedings{sima2024drivelm,
  title={Drivelm: Driving with graph visual question answering},
  author={Sima, Chonghao and Renz, Katrin and Chitta, Kashyap and Chen, Li and Zhang, Hanxue and Xie, Chengen and Bei{\ss}wenger, Jens and Luo, Ping and Geiger, Andreas and Li, Hongyang},
  booktitle={European conference on computer vision},
  pages={256--274},
  year={2024},
  organization={Springer}
}

@inproceedings{renz2025simlingo,
  title={Simlingo: Vision-only closed-loop autonomous driving with language-action alignment},
  author={Renz, Katrin and Chen, Long and Arani, Elahe and Sinavski, Oleg},
  booktitle={Proceedings of the Computer Vision and Pattern Recognition Conference},
  pages={11993--12003},
  year={2025}
}

@inproceedings{ding2023mitigating,
  title={Mitigating task interference in multi-task learning via explicit task routing with non-learnable primitives},
  author={Ding, Chuntao and Lu, Zhichao and Wang, Shangguang and Cheng, Ran and Boddeti, Vishnu Naresh},
  booktitle={Proceedings of the IEEE/CVF Conference on Computer Vision and Pattern Recognition},
  pages={7756--7765},
  year={2023}
}

@article{wang2024drivecot,
  title={Drivecot: Integrating chain-of-thought reasoning with end-to-end driving},
  author={Wang, Tianqi and Xie, Enze and Chu, Ruihang and Li, Zhenguo and Luo, Ping},
  journal={arXiv preprint arXiv:2403.16996},
  year={2024}
}

@article{jiang2024senna,
  title={Senna: Bridging large vision-language models and end-to-end autonomous driving},
  author={Jiang, Bo and Chen, Shaoyu and Liao, Bencheng and Zhang, Xingyu and Yin, Wei and Zhang, Qian and Huang, Chang and Liu, Wenyu and Wang, Xinggang},
  journal={arXiv preprint arXiv:2410.22313},
  year={2024}
}

@article{black2024pi_0,
  title={{$\pi_0$}: A Vision-Language-Action Flow Model for General Robot Control},
  author={Black, Kevin and Brown, Noah and Driess, Danny and Esmail, Adnan and Equi, Michael and Finn, Chelsea and Fusai, Niccolo and Groom, Lachy and Hausman, Karol and Ichter, Brian and others},
  journal={arXiv preprint arXiv:2410.24164},
  year={2024}
}

@inproceedings{li2026drive,
  title={Drive-r1: Bridging reasoning and planning in vlms for autonomous driving with reinforcement learning},
  author={Li, Yue and Tian, Meng and Zhu, Dechang and Zhu, Jiangtong and Lin, Zhenyu and Xiong, Zhiwei and Zhao, Xinhai},
  booktitle={Proceedings of the AAAI Conference on Artificial Intelligence},
  volume={40},
  number={8},
  pages={6708--6716},
  year={2026}
}

@article{jia2024bench2drive,
  title={Bench2drive: Towards multi-ability benchmarking of closed-loop end-to-end autonomous driving},
  author={Jia, Xiaosong and Yang, Zhenjie and Li, Qifeng and Zhang, Zhiyuan and Yan, Junchi},
  journal={Advances in Neural Information Processing Systems},
  volume={37},
  pages={819--844},
  year={2024}
}

@article{yu2020gradient,
  title={Gradient surgery for multi-task learning},
  author={Yu, Tianhe and Kumar, Saurabh and Gupta, Abhishek and Levine, Sergey and Hausman, Karol and Finn, Chelsea},
  journal={Advances in neural information processing systems},
  volume={33},
  pages={5824--5836},
  year={2020}
}

@article{zhao2025sce2drivex,
  title={Sce2drivex: A generalized mllm framework for scene-to-drive learning},
  author={Zhao, Rui and Yuan, Qirui and Li, Jinyu and Hu, Haofeng and Li, Yun and Gao, Zhenhai and Gao, Fei},
  journal={IEEE Robotics and Automation Letters},
  year={2025},
  publisher={IEEE}
}

@inproceedings{wu2022tcp,
  title={Trajectory-guided Control Prediction for End-to-end Autonomous Driving: A Simple yet Strong Baseline},
  author={Wu, Penghao and Jia, Xiaosong and Chen, Li and Yan, Junchi and Li, Hongyang and Qiao, Yu},
  booktitle={Advances in Neural Information Processing Systems},
  volume={35},
  year={2022}
}

@inproceedings{jia2023thinktwice,
  title={Think Twice before Driving: Towards Scalable Decoders for End-to-End Autonomous Driving},
  author={Jia, Xiaosong and Wu, Penghao and Chen, Li and Xie, Jiangwei and He, Conghui and Yan, Junchi and Li, Hongyang},
  booktitle={Proceedings of the IEEE/CVF Conference on Computer Vision and Pattern Recognition},
  year={2023}
}

@inproceedings{jia2023driveadapter,
  title={DriveAdapter: Breaking the Coupling Barrier of Perception and Planning in End-to-End Autonomous Driving},
  author={Jia, Xiaosong and Gao, Yulu and Chen, Li and Yan, Junchi and Liu, Patrick Langechuan and Li, Hongyang},
  booktitle={Proceedings of the IEEE/CVF International Conference on Computer Vision},
  year={2023}
}

@article{zhai2023admlp,
  title={Rethinking the Open-Loop Evaluation of End-to-End Autonomous Driving in nuScenes},
  author={Zhai, Jiang-Tian and Feng, Ze and Du, Jinhao and Mao, Yongqiang and Liu, Jiang-Jiang and Tan, Zichang and Zhang, Yifu and Ye, Xiaoqing and Wang, Jingdong},
  journal={arXiv preprint arXiv:2305.10430},
  year={2023}
}

@article{jiang2023vad,
  title={VAD: Vectorized Scene Representation for Efficient Autonomous Driving},
  author={Jiang, Bo and Chen, Shaoyu and Xu, Qing and Liao, Bencheng and Chen, Jiajie and Zhou, Helong and Zhang, Qian and Liu, Wenyu and Huang, Chang and Wang, Xinggang},
  journal={ICCV},
  year={2023}
}

@inproceedings{jia2025drivetransformer,
  title={DriveTransformer: Unified Transformer for Scalable End-to-End Autonomous Driving},
  author={Jia, Xiaosong and You, Junqi and Zhang, Zhiyuan and Yan, Junchi},
  booktitle={International Conference on Learning Representations},
  year={2025}
}

@inproceedings{fu2025orion,
  title={ORION: A Holistic End-to-End Autonomous Driving Framework by Vision-Language Instructed Action Generation},
  author={Fu, Haoyu and Zhang, Diankun and Zhao, Zongchuang and Cui, Jianfeng and Liang, Dingkang and Zhang, Chong and Zhang, Dingyuan and Xie, Hongwei and Wang, Bing and Bai, Xiang},
  booktitle={Proceedings of the IEEE/CVF International Conference on Computer Vision},
  pages={24823--24834},
  year={2025}
}

@article{zhou2025autovla,
  title={AutoVLA: A Vision-Language-Action Model for End-to-End Autonomous Driving with Adaptive Reasoning and Reinforcement Fine-Tuning},
  author={Zhou, Zewei and Cai, Tianhui and Zhao, Seth Z. and Zhang, Yun and Huang, Zhiyu and Zhou, Bolei and Ma, Jiaqi},
  journal={arXiv preprint arXiv:2506.13757},
  year={2025}
}

@article{xie2026flare,
  title={FLARE: Learning Future-Aware Latent Representations from Vision-Language Models for Autonomous Driving},
  author={Xie, Chengen and Sima, Chonghao and Li, Tianyu and Sun, Bin and Wu, Junjie and Hao, Zhihui and Li, Hongyang},
  journal={arXiv preprint arXiv:2601.05611},
  year={2026}
}

@article{li2025recogdrive,
  title={ReCogDrive: A Reinforced Cognitive Framework for End-to-End Autonomous Driving},
  author={Li, Yongkang and Xiong, Kaixin and Guo, Xiangyu and Li, Fang and Yan, Sixu and Xu, Gangwei and Zhou, Lijun and Chen, Long and Sun, Haiyang and Wang, Bing and Ma, Kun and Chen, Guang and Ye, Hangjun and Liu, Wenyu and Wang, Xinggang},
  journal={arXiv preprint arXiv:2506.08052},
  year={2025}
}

@article{li2025drivevlaw0,
  title={DriveVLA-W0: World Models Amplify Data Scaling Law in Autonomous Driving},
  author={Li, Yingyan and Shang, Shuyao and Liu, Weisong and Zhan, Bing and Wang, Haochen and Wang, Yuqi and Chen, Yuntao and Wang, Xiaoman and An, Yasong and Tang, Chufeng and Hou, Lu and Fan, Lue and Zhang, Zhaoxiang},
  journal={arXiv preprint arXiv:2510.12796},
  year={2025}
}

@article{yang2025drivemoe,
  title={DriveMoE: Mixture-of-Experts for Vision-Language-Action Model in End-to-End Autonomous Driving},
  author={Yang, Zhenjie and Chai, Yilin and Jia, Xiaosong and Li, Qifeng and Shao, Yuqian and Zhu, Xuekai and Su, Haisheng and Yan, Junchi},
  journal={arXiv preprint arXiv:2505.16278},
  year={2025}
}

\end{document}